\documentclass{article}

\usepackage{microtype}
\usepackage{graphicx}
\usepackage{subfigure}
\usepackage{booktabs}
\usepackage{hyperref}
\usepackage{float}

\usepackage[preprint]{icml2026}

\usepackage{amsmath}
\usepackage{amssymb}
\usepackage{mathtools}
\usepackage{amsthm}
\usepackage[capitalize,noabbrev]{cleveref}

\theoremstyle{plain}

\theoremstyle{definition}

\theoremstyle{remark}

\usepackage[textsize=tiny]{todonotes}

\icmltitlerunning{Late Interaction Retrieval for Protein Homolog Search}

\begin{document}

\twocolumn[
\icmltitle{\textsc{ProtoCol}: Late Interaction Retrieval for Protein Homolog Search}

\icmlsetsymbol{equal}{*}

\begin{icmlauthorlist}
\icmlauthor{Gabrielle Cohn}{equal,yyy}
\icmlauthor{Rohan Gumaste}{equal,yyy}
\icmlauthor{Minh Hoang}{equal,zzz}
\icmlauthor{Vihan Lakshman}{equal,yyy}
\end{icmlauthorlist}

\icmlaffiliation{yyy}{MIT}
\icmlaffiliation{zzz}{Princeton University}

\icmlcorrespondingauthor{Gabrielle Cohn}{gcohn@mit.edu}
\icmlcorrespondingauthor{Rohan Gumaste}{rgumaste@mit.edu}
\icmlcorrespondingauthor{Minh Hoang}{minhhoang@princeton.edu}
\icmlcorrespondingauthor{Vihan Lakshman}{vihan@mit.edu}

\icmlkeywords{Homolog Retrieval, Protein Representation Learning, Late Interaction}

\vskip 0.3in
]

\printAffiliationsAndNotice{\icmlEqualContribution} 

\begin{abstract}
Protein homology search underlies function annotation, structure prediction, and evolutionary analysis, but remains challenging in the ``twilight zone,'' where global sequence similarity is weak and classical alignment methods lose sensitivity. Protein language models provide context-aware representations that could improve alignment sensitivity in this regime. However, prior protein embedding-based retrieval pipelines often pool these representations into a single vector, potentially obscuring local motifs, domains, or conserved residues that reveal remote homology. We introduce \textsc{ProtoCol}, a model which represents proteins as sets of residue embeddings and uses ColBERT-style late interaction to test whether residue-level comparison improves homolog retrieval. \textsc{ProtoCol} encodes proteins independently, keeps candidate representations pre-computable, and scores candidates with MaxSim over residue embeddings. On SCOPe superfamily and Pfam clan benchmarks, \textsc{ProtoCol} outperforms sequence-composition, alignment-based, pooled PLM, and trained single-vector baselines, supporting late interaction as an effective retrieval layer for remote homology search.
\end{abstract}

\section{Introduction}

Homologous proteins descend from a common ancestral sequence and often preserve related functions or structures. Detecting such relationships is central to computational biology, supporting function annotation, structure prediction, and evolutionary analysis. This task is especially difficult for remote homologs, whose sequences may diverge so substantially that direct sequence-level similarity becomes weak.  In this ``twilight zone,'' classical sequence-alignment methods can miss relationships that remain evident through conserved motifs, domains, or structural constraints \cite{altschul1990basic, eddy2011accelerated, steinegger2017mmseqs2}.

Protein language models (PLMs) offer a promising sequence-only alternative because they produce contextual residue embeddings that encode structural and evolutionary signal \cite{lin2023evolutionary, Liu2024PLMSearch:}. However, many PLM-based retrieval pipelines pool these embeddings into a single protein vector and compare proteins by cosine similarity \cite{Iovino2024Protein}. This is efficient, but can dilute local evidence such as a conserved motif, domain, or small set of structurally constrained residues that may be decisive for remote homology.

This pooling bottleneck motivates our central question: \emph{does homolog retrieval improve when proteins are represented as sets of residue embeddings and compared through late interaction?} We hypothesize that residue-level scoring can preserve local evolutionary evidence while keeping database proteins independently encodable.


We test our hypothesis with \textsc{ProtoCol} (``proteins" with ``ColBERT") (see figure \ref{sec:a1}), a late-interaction retrieval model for protein homology search. \textsc{ProtoCol} adapts the ColBERT retrieval paradigm~\cite{khattab2020colbert} to protein sequences by representing each protein as residue-level PLM embeddings and comparing proteins through lightweight residue-level interactions\footnote{Our code is available at \url{https://github.com/gabriellecohn/ProtoCol}}. 


We evaluate \textsc{ProtoCol} on SCOPe superfamily and Pfam clan retrieval benchmarks using baselines chosen to isolate sequence composition, alignment sensitivity, PLM scale, contrastive fine-tuning, and late interaction. Across both settings, \textsc{ProtoCol} performs best, supporting residue-level late interaction as an effective retrieval layer for remote homology search.

\section{Related Work}

\paragraph{Alignment-based homology search.}
Classical homology search relies on residue-level sequence comparison. BLAST \cite{altschul1990basic} performs fast pairwise local alignment, while profile-based methods such as PSI-BLAST and HMMER \cite{eddy2011accelerated} improve sensitivity using multiple sequence alignments. MMseqs2 \cite{steinegger2017mmseqs2} further improves the speed--sensitivity tradeoff for large-scale search. These methods remain strong baselines, but their reliance on detectable sequence similarity limits sensitivity for highly diverged homologs.

Structure-based search addresses some of these limitations by comparing proteins closer to their conserved three-dimensional form. Foldseek \cite{vankempen2024fast}, for example, discretizes protein structures into residue-level alphabets and aligns the resulting strings. Such methods demonstrate the value of local matching beyond raw sequence identity, but require structural information to work. Our work instead asks whether sequence-only PLM representations can support an analogous retrieval primitive.

\textbf{Protein language models for retrieval.}
Large protein language models such as ProtTrans \cite{elnaggar2021prottrans} and ESM \cite{lin2023evolutionary} learn contextual residue embeddings from unaligned sequence databases. These embeddings have been used for retrieval via mean-pooling into a single protein-level vector and ranking candidates by cosine similarity \cite{10.3389/fbinf.2022.1033775, Iovino2024Protein}. This bi-encoder-style setup supports efficient nearest-neighbor search, analogous to dense retrieval methods in NLP such as Sentence-BERT \cite{reimers2019sentence} and Dense Passage Retrieval \cite{karpukhin2020dense}, but discards the token-level structure of PLM embeddings. Methods such as PLMAlign and pLM-BLAST \cite{Liu2024PLMSearch:, 10.1093/bioinformatics/btad579} point to the utility of residue-level PLM embeddings for remote homology detection by using PLM representations in alignment-oriented comparison.

\textbf{Late-interaction retrieval.}
Late-interaction models occupy a middle ground between efficient global-vector retrieval and expensive pairwise interaction models. ColBERT \cite{khattab2020colbert} represents each text passage as token embeddings and scores a query--document pair by matching each query token to its most similar document token using MaxSim. This preserves fine-grained matching while allowing document representations to be precomputed. Following the introduction of ColBERT, a number of works have further developed this notion of late interaction retrieval~\cite{santhanam2022colbertv2, santhanam2022plaid, formal2024splate, faysse2025colpali, lee2023rethinking, lin2023fine, chaffin2025pylate, dhulipala2024muvera, engels2023dessert}. All of these prior works focus on traditional information retrieval settings. \textsc{ProtoCol} adapts this idea to proteins: residues replace tokens, candidate proteins replace documents, and MaxSim provides a residue-level similarity score for homolog retrieval. In contrast to alignment-oriented PLM methods, \textsc{ProtoCol} uses late interaction as a learned retrieval mechanism rather than computing an explicit alignment path.

\section{Methodology}
\label{sec:method}

This section defines the components of \textsc{ProtoCol}: a residue-level PLM encoder, MaxSim scoring, and a contrastive objective based on weak homology labels. Together, these choices test whether homology retrieval benefits from preserving residue embeddings through the scoring layer rather than compressing each protein before comparison. Dataset construction, baselines, and evaluation protocol are described in \Cref{sec:experiments}.

\textbf{Late-interaction encoder.}
We instantiate ColBERT-style late interaction over a protein language model. Let $x = (x_1, \dots, x_T)$ be a protein sequence, and let $h_t \in \mathbb{R}^H$ denote the contextual residue embedding produced at position $t$ by an ESM-2 backbone $f_\theta$ \cite{lin2023evolutionary}. We attach a linear projection $W \in \mathbb{R}^{D \times H}$ followed by L2 normalization:
\begin{equation}
    e_t = \frac{W h_t}{\lVert W h_t \rVert_2} \in \mathbb{R}^D,
\end{equation}
with $D = 128$ in all experiments. Unless otherwise stated, the backbone is ESM-2 35M ($H=480$, 12 layers). To keep the trainable footprint modest, we freeze the embedding layer and lower transformer blocks and fine-tune only the final three transformer layers, the post-stack LayerNorm, and $W$. This yields roughly 8.4M trainable parameters out of 33.6M.

\textbf{MaxSim scoring.} A protein is represented by its variable-length set of L2-normalized residue embeddings $E = \{e_1, \dots, e_T\}$. MaxSim operationalizes residue-level retrieval by allowing each query residue to contribute its strongest match anywhere in the candidate protein. Given query embeddings $E^q$ and candidate embeddings $E^d$, we score the pair using the asymmetric MaxSim operator of \citet{khattab2020colbert},
\begin{equation}
    \mathrm{MaxSim}(E^q, E^d)
    = \sum_{i=1}^{T_q} \max_{j \in [T_d]} \langle e^q_i, e^d_j \rangle .
\end{equation}
Because embeddings are L2-normalized, each inner product is a cosine similarity. Padding positions are masked in both the inner maximum and outer sum.

\textbf{Contrastive training.}
Training shapes the embedding space so homologs receive high late-interaction scores and in-batch non-homologs receive lower scores. Each training pair consists of an anchor protein $a$ and a positive protein $p$ sampled from the same superfamily for SCOPe~\cite{chandonia2022scope} or from the same clan for Pfam~\cite{10.1093/nar/gkaa913}. For a batch of $B$ pairs, we form $S \in \mathbb{R}^{B \times B}$ with $S_{ij} = \mathrm{MaxSim}(E^{a_i}, E^{p_j})$, treat off-diagonal entries as in-batch negatives, and minimize the symmetric InfoNCE objective
\begin{equation}
    \mathcal{L}
    = \tfrac{1}{2}\!\left[
    \mathrm{CE}(S/\tau, y) + \mathrm{CE}(S^\top/\tau, y)
    \right],
\end{equation}
where $y_i=i$ and $\tau$ is a temperature. We do not filter accidental positive collisions among in-batch negatives.

\paragraph{Implementation details.}
Sequences are tokenized with the ESM-2 tokenizer and truncated to $T \leq 256$ residues. We optimize with AdamW using weight decay $0.01$ for three epochs at batch size 16. The learning rate follows a OneCycleLR schedule with peak learning rate $2 \times 10^{-5}$ and 10\% warmup. Training uses fp16 autocast on a single GPU; gradients are unscaled and clipped to global norm 1.0 before each step. We set $\tau=1$ throughout.

\section{Experiments}
\label{sec:experiments}


\subsection{Datasets and Retrieval Protocol}
\label{sec:data}

We evaluate homolog retrieval in two complementary settings. SCOPe provides a hierarchical structural classification of protein domains; we use superfamily labels as evidence of shared ancestry among potentially remote homologs \cite{chandonia2022scope}. Pfam groups protein sequences into families using sequence alignments and profile HMMs; we use clan labels, which group related families, as a broader test of remote homology retrieval.

For each dataset, we construct train and test splits over evolutionary groups and train a separate \textsc{ProtoCol} model on that dataset's training split. At evaluation time, each protein is used as a query against the remaining proteins in the corresponding test database, excluding self-matches. Retrieved proteins are relevant if they share the query's held-out evolutionary group: superfamily for SCOPe and Pfam clan for Pfam. Group-disjoint train/test splits ensure that evaluation measures generalization to unseen homologous groups rather than memorization of training labels.

\subsection{Compared Methods}
\label{sec:baselines}

We compare against baselines designed to distinguish the contribution of late-interaction scoring from other sources of retrieval signal: sequence composition, alignment sensitivity, PLM scale, contrastive fine-tuning, and pretrained residue similarity without task-specific adaptation. 
The trained \textsc{ProtoCol} model is described in \Cref{sec:method}.

\textbf{MinHash Jaccard.}
We compute MinHash-approximated Jaccard similarity over amino acid 5-mers. Each sequence is decomposed into overlapping 5-mers, and a MinHash signature with 256 permutations is computed using \texttt{datasketch}. Candidates are ranked by the fraction of matching hash values between signatures.

\textbf{MMseqs2.}
We evaluate MMseqs2 \cite{steinegger2017mmseqs2} as a strong alignment-based sequence retrieval baseline. All test sequences are searched against the full test set using \texttt{mmseqs easy-search} with sensitivity 7.5. Hits are ranked by e-value in ascending order.

\textbf{Mean-pooled ESM-2 650M.}
To assess the importance of encoder scale, we embed each protein using frozen ESM-2 650M (\texttt{facebook/esm2\_t33\_650M\_UR50D}), which is substantially larger than the ESM-2 35M backbone used by \textsc{ProtoCol}. Final-layer residue embeddings are mean-pooled, L2-normalized, and ranked by cosine similarity.

\textbf{Uni-vector ESM-2 35M.}
This is the direct ablation of late interaction. It uses the same ESM-2 35M backbone and contrastive objective as \textsc{ProtoCol}, but mean-pools residue embeddings into one L2-normalized protein vector and retrieves by cosine similarity. This tests whether gains come from residue-level scoring rather than fine-tuning alone.

\textbf{Frozen \textsc{ProtoCol}.}
To isolate task-specific optimization, we evaluate a frozen \textsc{ProtoCol} variant with the same ESM-2 35M backbone, 128-dimensional projection, and MaxSim scoring function as the trained model, but with all parameters left at their initial values.

\subsection{Evaluation Metric}
\label{sec:metrics}
We evaluate retrieval using the capped recall@k metric,
\begin{equation}
\mathrm{cRecall@}k(q) = \frac{\mathrm{hits@}k(q)}{\min(k, N_q)},
\end{equation}
where $\mathrm{hits@}k(q)$ is the number of top-$k$ retrieved proteins that are real homologs of $q$, and $N_q$ is the number of other proteins in that group. Unlike standard recall@$k$, capped recall@$k$ normalizes by the maximum number of relevant proteins that can appear in the top $k$, so a ranking whose top-$k$ entries are all relevant receives a score of 1 regardless of group size \cite{ji2025targetbenchmarkingtableretrieval, chen-etal-2023-breaking}. We computed capped recall by treating every protein in the test set as a query and retrieving from the remaining set of examples.  

\subsection{Results}
\label{sec:results}

\begin{table}[t]
\centering
\small
\setlength{\tabcolsep}{3.5pt}
\begin{tabular}{lcccc}
\toprule
Method & cR@1 & cR@10 & cR@100 & Lat. (ms) \\
\midrule
MinHash            & 0.3172 & 0.1932 & 0.1567 & \textbf{4.23} \\
MMseqs2            & 0.8578 & 0.6177 & 0.3421 & 1068.73 \\
ESM-2 650M pool    & 0.8773 & 0.6664 & 0.5127 & 49.58 \\
ESM-2 35M uni      & 0.6819 & 0.6777 & 0.7141 & 20.31 \\
\textsc{ProtoCol}-F & 0.8825 & 0.8093 & 0.7422 & 24.33 \\
\textbf{\textsc{ProtoCol}} & \textbf{0.9460} & \textbf{0.8947} & \textbf{0.8796} & 24.10 \\
\bottomrule
\end{tabular}
\caption{SCOPe superfamily retrieval performance ($n_{\mathrm{test}}=2314$). Scores are capped recall; latency is in ms. MMseqs2 uses sensitivity 7.5; MinHash uses 5-mers; \textsc{ProtoCol}-F is frozen.}
\label{tab:scope_results}
\vspace{-5mm}
\end{table}

\begin{table}[t]
\centering
\small
\setlength{\tabcolsep}{4pt}
\begin{tabular}{lcccc}
\toprule
Method & cR@1 & cR@10 & cR@100 & Lat. (ms) \\
\midrule
MinHash             & 0.4957 & 0.1684 & 0.1104 & \textbf{3.54} \\
MMseqs2             & 0.8833 & 0.3627 & 0.1539 & 1050.74 \\
ESM-2 650M pool     & 0.8903 & 0.6048 & 0.4324 & 49.86 \\
ESM-2 35M uni       & 0.7897 & 0.6657 & 0.6378 & 19.14 \\
\textsc{ProtoCol}-F & 0.8147 & 0.5881 & 0.4533 & 24.90 \\
\textbf{\textsc{ProtoCol}} & \textbf{0.9377} & \textbf{0.7630} & \textbf{0.7065} & 26.49 \\
\bottomrule
\end{tabular}
\caption{Pfam clan retrieval performance ($n_{\mathrm{test}}=3000$). Metrics and parameter settings are identical to Table \ref{tab:scope_results}.}
\label{tab:pfam_results}
\vspace{-8mm}
\end{table}
\textbf{\textsc{ProtoCol} outperforms other retrieval baselines.} Across both benchmarks, trained \textsc{ProtoCol} achieves the strongest performance at every cutoff. On SCOPe, it improves over the next best baseline by 6.87 points at cR@1, 21.07 at cR@10, and 16.55 at cR@100. These gains are largest at deeper retrieval cutoffs, suggesting that ranking many homologs requires evidence beyond a single global representation or sequence-level similarity comparison.

The Uni-vector ESM-2 35M baseline is the key ablation. It uses the same backbone and supervision as \textsc{ProtoCol}, but removes residue-level scoring. Because this comparison holds the encoder and training signal fixed, \textsc{ProtoCol}'s consistent gains indicate that late interaction adds value beyond contrastive fine-tuning alone.

\begin{figure}[t!]
\centering
\includegraphics[width=\columnwidth]{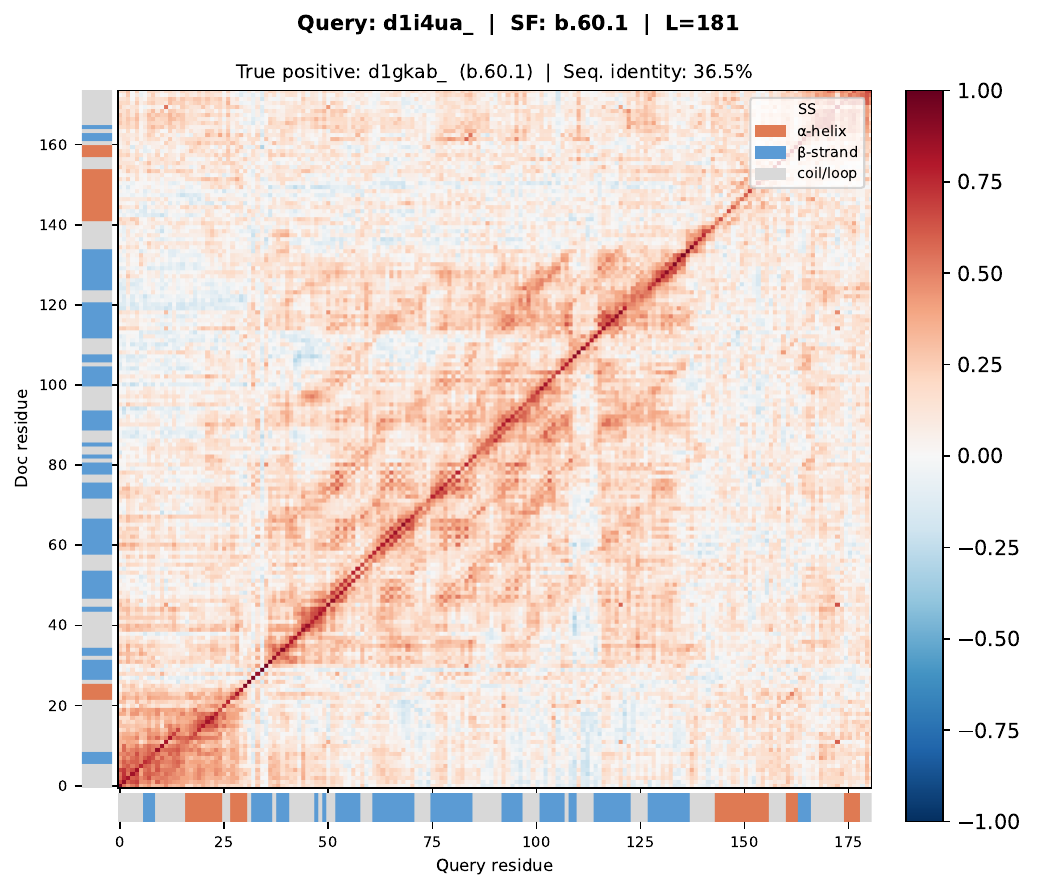}
\caption{
\textbf{ColBERT attention maps for true positive pair.} We visualize the residue-level similarity matrix between a representative query and its highest-ranked true positive match. Secondary structure annotations are shown along each axis. The similarity map exhibits block diagonal structure that coincides with secondary structure boundaries, indicating that \textsc{ProtoCol} indeed learns meaningful structural organization patterns to facilitate retrieval.
}
\label{fig:attention-map}
\vspace{-5mm}
\end{figure}

Frozen \textsc{ProtoCol} further separates the contribution of architecture from task-specific training. Its competitive SCOPe performance suggests that pretrained ESM-2 residue embeddings already encode useful local similarity structure. However, the substantial gains of the trained model indicate that weak homology supervision sharpens this structure further — as evidenced by the block-diagonal similarity maps in Fig.~\ref{fig:attention-map}, where the trained model produces coherent high-similarity blocks that align with secondary structure boundaries, suggesting that contrastive fine-tuning encourages the embeddings to organize similarity along structurally meaningful lines rather than purely sequential ones.

The Pfam benchmark shows the same overall pattern, with a sharper distinction between nearest-neighbor and deeper retrieval. MMseqs2 and frozen mean-pooled ESM-2 650M remain competitive at cR@1, but drop more substantially at cR@10 and cR@100. In contrast, trained \textsc{ProtoCol} improves over trained Uni-vector ESM-2 35M by 14.80, 9.73, and 6.87 points at cR@1, cR@10, and cR@100, respectively. This supports the benefit of residue-level late interaction under a different protein-family taxonomy.

\textbf{\textsc{ProtoCol} embeddings reflect structural organization.} We investigate 
whether the retrieval mechanism learned by \textsc{ProtoCol} is tied to the biological 
structure of protein sequences. We obtain PDB structure files for all SCOPe domain 
sequences in the test set and annotate each domain with per-residue secondary structure 
labels using the pyDSSP package. The DSSP algorithm assigns each residue one of three 
coarse-grained secondary structure states: $\alpha$-helix (orange), $\beta$-strand 
(blue), or coil/loop (gray). The annotations for an exemplary query and its top-ranked 
true positive retrieval are overlaid along both axes of their pairwise residue embedding 
similarity matrix (Fig.~\ref{fig:attention-map}).

Transitions between secondary structure elements, particularly between $\beta$-strands and coil/loop regions, correspond to visible discontinuities in the similarity map. At these boundaries, similarity drops sharply, producing a block-diagonal appearance that corresponds to shared secondary structure elements. For example, the $\beta$-strand-dominated region spanning residues 35-135 on the query and residues 25--130 on the top-hit forms a large, coherent diagonal block. This finding provides evidence that \textsc{ProtoCol}'s learned embeddings implicitly encode structural organization beyond raw sequence identity.

\section{Conclusion}
\label{sec:conclusion}
Taken together, these comparisons show that representing proteins as sets of residue 
embeddings and comparing them through late interaction improves homolog retrieval. The 
strongest controlled evidence comes from the trained Uni-vector comparison, which holds 
the backbone and supervision fixed while replacing residue-level MaxSim scoring with a 
single global vector. \textsc{ProtoCol}'s consistent gains show that the benefit is not 
only due to contrastive fine-tuning, but also to preserving and comparing local 
residue-level evidence. The substantial gains of the trained model over the frozen variant indicate 
that weak homology supervision sharpens understanding of secondary structure, which is likely responsible for the improved performance.

In future, we hope to build on these preliminary results and, in particular, investigate how to scale \textsc{ProtoCol} to perform efficient late-interaction search over orders-of-magnitude larger protein databases. 


\bibliographystyle{icml2026}
\bibliography{references}

\newpage
\clearpage
\appendix 
\section{Appendix}
\label{sec:appendix}
\subsection{Framework Overview}
\label{sec:a1}

\begin{figure}[H]
    \centering
    \begin{minipage}{0.7\textwidth}
        \includegraphics[
            width=\textwidth,
            height=\textheight,
            keepaspectratio
        ]{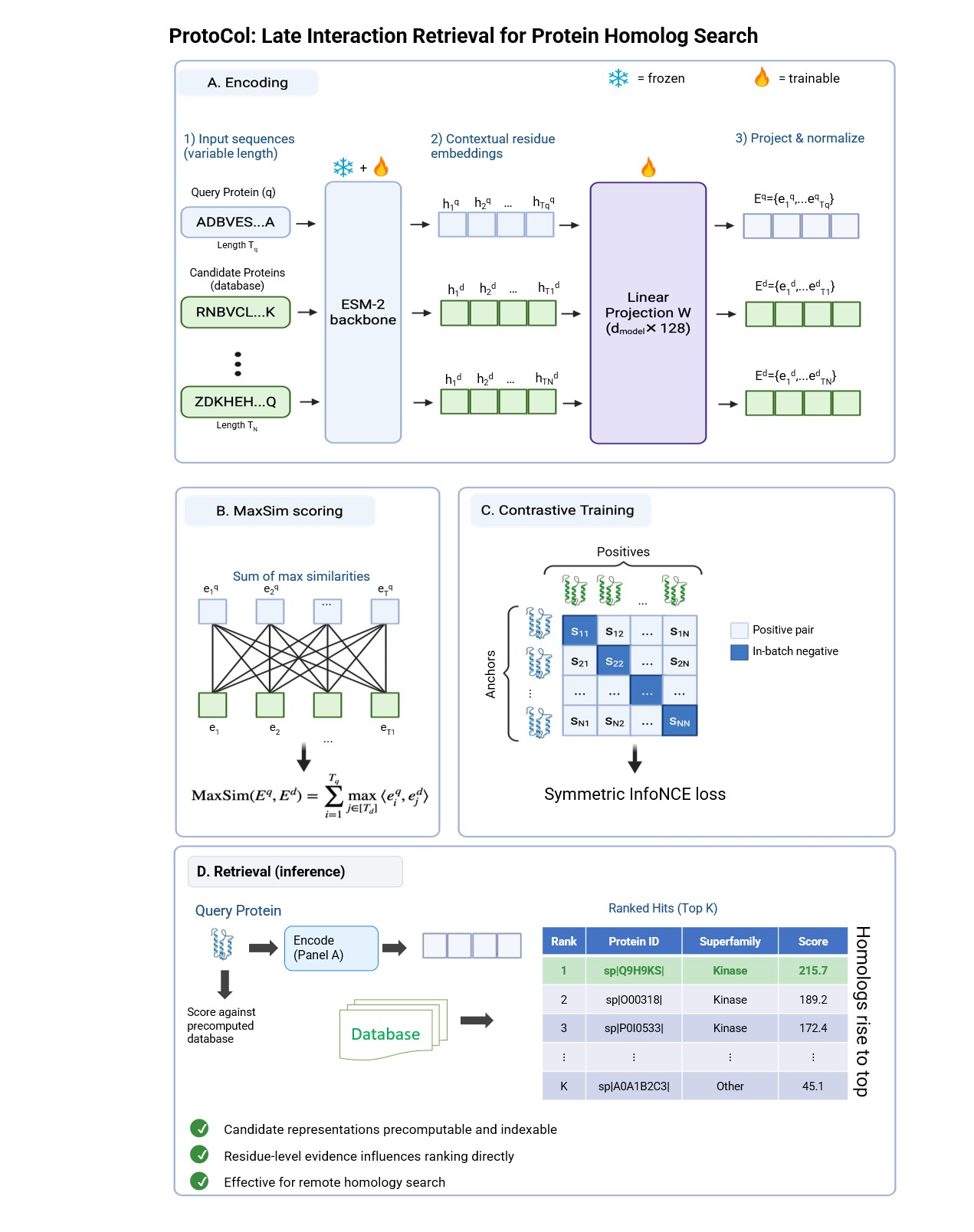}
        \caption{Overview of the ProtoCoL framework for protein homolog retrieval. Variable-length query and candidate protein sequences are encoded with a frozen ESM-2 backbone, projected into residue-level embeddings, and compared using MaxSim scoring. The projection layer is trained with a symmetric contrastive objective, enabling retrieval using precomputed candidate representations.}
        \label{fig:intro_example}
    \end{minipage}
\end{figure}

\end{document}